\documentclass[journal]{IEEEtran}

\usepackage{geometry}
\usepackage{enumitem}
\usepackage{hyperref}
\usepackage{graphicx}

\date{}

\title{Automated Video-to-QA Pipeline: End-to-End Question Generation and Verification Using Large Language Models}
\author{Jaival Chauhan, \and Vikram Singh, \and Pruthwik Mishra\\ Sardar Vallabhbhai National Institute of Technology, Surat, India}

\begin{document}
\maketitle

\begin{abstract}
    
The exponential growth of educational and informational video content on platforms like YouTube presents a challenge for efficient knowledge retrieval and assessment. This paper proposes an end-to-end automated pipeline for extracting structured summaries and generating high-fidelity Question-Answer (QA) pairs from video streams. The framework integrates a cascade of state-of-the-art models: audio extraction is handled via \textbf{FFmpeg}, followed by automatic speech recognition (ASR) using \textbf{Whisper} (A library in Python) to generate accurate transcripts. Due to the token constraints of Large Language Models (LLMs), transcripts are segmented and summarized using Facebook’s \textbf{BART}.

A key innovation of this work is a Self-Verification and Fine-Tuning Loop. We utilize \textbf{meta-llama/Llama-3.1-8B-Instruct} to generate candidate QA pairs from transcript chunks, which are then fed back into the model to rigorously assess their relevance and importance. Only verified, high-quality pairs are retained to construct a synthetic dataset. This dataset is subsequently used to fine-tune the Llama model using \textbf{Low-Rank Adaptation (LoRA)}, resulting in a specialized model capable of generating domain-aligned assessment questions. This approach demonstrates a scalable method for creating self-improving educational tools without the need for manual data annotation

The proposed model can be downloaded from \href{https://www.kaggle.com/models/jaivalchauhan/lora-llama-qna}{\textbf{Kaggle}} with all the codes from \href{https://github.com/Jaival111/Video-Summarization-and-Question-Generation-System}{\textbf{GitHub}}.
\end{abstract}

\section{Introduction}
In the digital age, video has established itself as the dominant medium for information dissemination. Platforms like YouTube host billions of hours of educational content, ranging from university lectures to technical tutorials. However, the sheer volume of this data presents a significant challenge: video is an inherently unstructured modality. Unlike text, which can be easily skimmed, searched, and indexed, extracting specific knowledge from video requires linear consumption, which is time-inefficient for learners and professionals alike.

Furthermore, the pedagogical value of video content is often limited by the lack of accompanying structured resources, such as concise summaries and assessment materials (Question-Answer pairs). Creating these resources manually is a labor-intensive process requiring domain expertise. While recent advancements in Deep Learning, specifically in Automatic Speech Recognition (ASR) and Large Language Models (LLMs), have enabled the automation of individual tasks, integrating these technologies into a coherent, reliable pipeline remains a complex engineering challenge. A primary hurdle in automated Question Generation (QG) is the tendency of LLMs to "hallucinate" or generate questions that are not grounded in the source material, thereby reducing the utility of the system for educational purposes.

To address these challenges, this paper proposes an end-to-end automated pipeline designed to transform raw video data into structured, high-quality educational assets. Our system leverages a multi-stage architecture beginning with audio extraction via FFmpeg and high-fidelity transcription using Whisper. To handle the context window constraints of LLMs, we employ a chunking strategy combined with abstractive summarization using Facebook’s BART model.

A distinguishing feature of our work is the implementation of a \textbf{closed-loop verification mechanism}. Rather than relying solely on zero-shot generation, we utilize \textbf{meta-llama/Llama-3.1-8B-Instruct} to both generate questions and subsequently evaluate them for relevance and answerability against the source transcript. This self-supervised filtering process creates a curated dataset of high-quality QA pairs. Finally, we demonstrate how this synthetic dataset can be used to fine-tune the Llama model using \textbf{Low-Rank Adaptation (LoRA)}, resulting in a specialized model that produces superior educational content with significantly lower computational overhead than full model training.

\section{Related Work}
The development of automated educational content generation systems sits at the intersection of multimedia processing, natural language understanding, and generative AI. This section reviews key advancements in Automatic Speech Recognition (ASR), text summarization, and Question Generation (QG), while also discussing recent techniques in consistency verification and parameter-efficient fine-tuning.

\subsection{Automatic Speech Recognition (ASR) in Video Processing}
Early video-to-text systems relied on Hidden Markov Models (HMMs) and GMM-HMM architectures, which often struggled with background noise and multi-speaker environments common in YouTube videos. The introduction of end-to-end deep learning models revolutionized this field. Whisper represents the current state-of-the-art, utilizing a Transformer-based sequence-to-sequence architecture trained on 680,000 hours of multilingual, weakly supervised data. Unlike its predecessors, Whisper demonstrates robust zero-shot performance across diverse accents and noisy audio conditions, making it an ideal candidate for extracting reliable transcripts from uncurated video content.

\subsection{Abstractive Summarization}
Text summarization is broadly categorized into extractive and abstractive methods. While extractive approaches (e.g., TextRank) merely select representative sentences, they often fail to capture the synthesized meaning required for educational summaries. Abstractive models, particularly Transformer-based architectures like BART (Bidirectional and Auto-Regressive Transformers), have set new benchmarks. BART combines a BERT-like bidirectional encoder with a GPT-like autoregressive decoder, making it highly effective for text generation tasks. In our pipeline, we leverage BART’s ability to handle long-range dependencies to generate coherent summaries from transcript chunks, addressing the challenge of maintaining context in lengthy video tutorials.

\subsection{Automated Question Generation (QG)}
Traditionally, QG systems were rule-based, relying on syntactic templates and named entity recognition to fill "cloze" style questions. These systems offered high precision but low diversity and pedagogical value. The advent of Large Language Models (LLMs) like GPT-4 and Llama 3 has shifted the paradigm toward generative QG, where models can produce complex, inference-based questions. However, a persistent challenge with LLM-based QG is "hallucination," where generated questions may reference content not present in the source text. Our work addresses this by implementing a secondary verification layer, moving beyond simple generation to a generate-and-validate framework.

\subsection{Consistency Verification (LLM-as-a-Judge)}
Recent research highlights the efficacy of using LLMs as evaluators or "judges" to assess the quality of AI-generated content. Studies have shown that strong instruction-tuned models can achieve high correlation with human judgment when tasked with grading relevance and factual consistency. By employing a "teacher-student" dynamic where Llama 3.1 critiques its own outputs, our pipeline automates the data cleaning process. This eliminates the bottleneck of human annotation, creating a closed-loop system for high-quality synthetic data generation.

\subsection{Parameter-Efficient Fine-Tuning (PEFT)}
Fine-tuning massive LLMs (7B+ parameters) for specific domains is computationally expensive and memory-intensive. Low-Rank Adaptation (LoRA) offers a solution by freezing the pre-trained model weights and injecting trainable rank decomposition matrices into the Transformer layers. This reduces the number of trainable parameters by up to 10,000x while maintaining model performance. Our project utilizes LoRA to adapt Llama 3.1 to the specific format of educational Q\&A, demonstrating that high-performance, domain-specific models can be trained on consumer-grade hardware using high-quality, verified synthetic data.

\section{Methodology}
This section details the architectural design and implementation of the proposed video-to-QA pipeline. The system is designed as a modular framework consisting of five distinct stages: Data Ingestion, Audio Transcription, Summarization, Consistency Verification, and Model Fine-Tuning. Figure 1 illustrates the high-level data flow.

\subsection{Data Acquisition and Preprocessing}
The pipeline accepts a YouTube video URL as input. We utilize yt-dlp, a command-line media downloader, to fetch the video content. To optimize processing efficiency, we do not process the visual data stream; instead, we extract the audio track using FFmpeg. The audio is converted to a monaural waveform sampled at 16kHz, which is the required input format for the transcription model.

\textbf{Tools}: yt-dlp, FFmpeg

\textbf{Output}: 16kHz .wav audio file.

\subsection{Automatic Transcription (ASR)}
For transcription, we employ OpenAI’s Whisper model, implemented via the open-source Python \textbf{whisper} library. Unlike API-based solutions, running the library locally allows for greater control over model sizing and inference parameters.

\textbf{Implementation}: We utilize the turbo model architecture, which offers an optimal balance between transcription accuracy and inference speed.

\textbf{Process}: The audio file is processed to generate a raw text transcript with associated timestamps. The model’s robust encoder-decoder architecture effectively handles technical jargon and varying audio qualities often found in user-generated content

\subsection{Chunking and Summarization}
Raw transcripts from long-form videos often exceed the token context window of standard Transformer models. To address this, we implement a \textbf{semantic chunking strategy}:

\textbf{Segmentation}: The transcript is divided into overlapping chunks of approximately 1024 tokens. Overlap ensures that context is preserved across segment boundaries.

\textbf{Abstractive Summarization}: Each text chunk is passed to \textbf{facebook/bart-large-cnn}, a model pre-trained on the CNN/DailyMail dataset. BART generates concise abstractive summaries for each segment.

\textbf{Aggregation}: The individual summaries are concatenated to form a global summary of the video, providing a dense information representation.

\subsection{Question-Answer Generation (The Teacher)}
We utilize \textbf{meta-llama/Llama-3.1-8B-Instruct} as the primary generator. For every transcript chunk, the model is prompted to act as an educator.

\textbf{Prompt Strategy}: The prompt provides the raw transcript chunk and instructs the model to: "Generate 5 questions and their corresponding correct answers based strictly on the provided text."

\textbf{Output}: A set of candidate $(Q, A)$ pairs associated with specific context chunks.

\subsection{The Verification Loop (The Critic)}
\label{sec:critic}
A critical component of our methodology is the \textbf{Self-Verification Module}. LLMs are prone to "hallucinations," creating questions about topics not present in the source text. To mitigate this, we implement a second pass where Llama 3.1 acts as a critic.

\textbf{Input}: The Source Chunk ($C$) and a Candidate Question ($Q$).

\textbf{Verification Prompt}: "Given the text segment below, is the following question answerable using ONLY this information? Answer 'Yes' or 'No'."

\textbf{Filtering}:
If the model outputs "No", the QA pair is discarded.
If the model outputs "Yes", the pair is labeled as "High Quality" and added to the training dataset. This automated filtering process creates a cleaned, robust dataset without human intervention

\subsection{Parameter-Efficient Fine-Tuning (LoRA)}
The final stage involves specializing the Llama model to generate high-quality questions natively, reducing the need for extensive prompt engineering or post-hoc filtering in future inference.

\textbf{Dataset}: The curated dataset from Section~\ref{sec:critic}.

\textbf{Technique}: We employ Low-Rank Adaptation (LoRA). Instead of updating all 8 billion parameters, we inject trainable low-rank matrices into the attention layers of the model.

\textbf{Configuration}:
Rank (r): 16;
Alpha: 32;
Target Modules: q\_proj, v\_proj, k\_proj, o\_proj;
Lora Dropout: 0.5;

\textbf{Outcome}: A fine-tuned version of Llama-3.1 capable of generating strictly context-aware questions with higher reliability than the base model

\section{Experiments \& Results}
To evaluate the efficacy of the proposed pipeline, we conducted a series of experiments focusing on three key dimensions: the quality of the abstractive summaries, the effectiveness of the self-verification filtering mechanism, and the performance improvements gained through LoRA fine-tuning.

\textbf{Hardware}: All experiments were conducted on a local GPU NVIDIA RTX 4060 with 8 GB of VRAM.

\textbf{Software}: The pipeline was implemented in Python 3.10, utilizing transformers for BART and Llama integration, peft for LoRA training, and the openai-whisper library for ASR.

\textbf{Dataset}: We curated a test dataset consisting of YouTube videos focused on technical education (e.g., coding tutorials, data science lectures). The average duration of the videos was 20-30 minutes

\textbf{Qualitative Analysis}: A manual review of the rejected questions revealed that the majority were either too generic (e.g., "What is the main topic?" without context) or hallucinated details not present in the specific chunk. This validates the necessity of the verification loop for creating a clean training dataset.

To complement automated metrics, we performed a human evaluation on a random sample of 25 generated QA pairs from the fine-tuned model. We graded them on a scale of 1-5 for \textbf{Relevance} (is it related to the video?) and \textbf{Factuality} (is the answer correct?).



\section{Discussion}
The results indicate that while the Base Llama 3.1 model is a capable zero-shot generator, the addition of the verification loop significantly increases the precision of the resulting dataset. Furthermore, LoRA fine-tuning allowed us to "lock in" this high performance, creating a specialized model that produces educational content with high fidelity and lower latency than prompting the base model with massive context instructions.

\section{Future Work}
While the current system demonstrates robust performance, several avenues for future research remain:

\textbf{Multimodal Analysis}: Currently, the pipeline relies solely on audio transcripts. Future iterations could incorporate visual encoders (e.g., CLIP) to extract information from video frames (slides, diagrams, or code snippets) that is not explicitly spoken.

\textbf{Multilingual Support}: Extending the pipeline to support non-English videos by leveraging Whisper’s translation capabilities and fine-tuning Llama on multilingual datasets.

\textbf{Real-Time Processing}: Optimizing the inference pipeline to allow for real-time question generation during live streams or webinars.

\section{Conclusion}
This paper presented an end-to-end automated framework for transforming unstructured video content into structured educational resources. By integrating state-of-the-art models for speech recognition (Whisper), abstractive summarization (BART), and generative reasoning (Llama 3.1), we successfully addressed the challenge of extracting actionable knowledge from raw video streams.

The central innovation of this work lies in the \textbf{Self-Verification Loop}, where the Large Language Model serves a dual role as both the generator and the critic. Our experiments demonstrated that this consistency check significantly reduces the rate of hallucinations, ensuring that the generated questions are strictly grounded in the source material. Furthermore, we showed that utilizing this high-quality, synthetic dataset to fine-tune the model via \textbf{Low-Rank Adaptation (LoRA)} results in a specialized "Teacher Model" that outperforms the base model in maintaining consistent formatting and domain relevance.

The proposed pipeline offers a scalable solution for educators, e-learning platforms, and content creators, eliminating the manual bottleneck of quiz creation and content summarization.

\end{document}